Introduction and Assessment of the Addition of Links and Containers to the Blackboard Architecture

Jordan Milbrath & Jeremy Straub
Institute for Cyber Security Education and Research
North Dakota State University
1320 Albrecht Blvd., Room 258
Fargo, ND 58108
Phone: +1-701-231-8196
Fax: +1-701-231-8255
Email: jordan.milbrath@ndsu.edu, jeremy.straub@ndsu.edu

**Abstract**

The Blackboard Architecture provides a mechanism for storing data and logic and using it to make decisions that impact the application environment that the Blackboard Architecture network models. While rule-fact-action networks can represent numerous types of data, the relationships that can be easily modeled are limited by the propositional logic nature of the rule-fact network structure.  This paper proposes and evaluates the inclusion of containers and links in the Blackboard Architecture. These objects are designed to allow them to model organizational, physical, spatial and other relationships that cannot be readily or efficiently implemented as Boolean logic rules.  Containers group related facts together and can be nested to implement complex relationships.  Links interconnect containers that have a relationship that is relevant to their organizational purpose.  Both objects, together, facilitate new ways of using the Blackboard Architecture and enable or simply its use for complex tasks that have multiple types of relationships that need to be considered during operations.

**1. Introduction**

The Blackboard Architecture [1] provides, conceptually, a mechanism for modeling both knowledge and decision-making.  Through this, it is able to understand and make decisions affecting an application domain.

There is no conceptual limit to the scope of a Blackboard Architecture, for a given domain, or even a requirement that a given system store and interact with only a single application domain. However, there is an inherent organizational limitation of rule-fact networks. Facts and rules can be used to create complex networks. However, as the networks grow in scale and the quantity of rules grow, so does the complexity (and, potentially, the incomprehensibility) of these networks. As more connections are created, it becomes increasingly difficult to organize and visualize the networks coherently.

Additionally, while the Blackboard Architecture model allows the representation of virtually any type of relationships between facts, rules are the only mechanism to implement these relationships and organization. If facts within a Blackboard Architecture network are related by multiple criteria, rules must be used for multiple purposes, potentially causing confusion. If they are related by any criteria that is not simply a logical association between their values, which can be modeled as a rule, there is no mechanism to model this.

Some applications benefit from the ability to group knowledge elements by what they represent, rather than their values' interactions. This separation of operational logic and other organizational relationships has the potential to provide multiple benefits, such as facilitating the replication of rule-fact logic for multiple instances of a given object type and the ability to model organizational, physical, spatial, and other relationships which may be relevant to system operations and decision making. While these relationships could be modeled as complex interaction rule-sets, this would be difficult to understand, cumbersome to update, and increase the size of the operational network significantly.

Disorganization within blackboard networks can result in reduced usability, ability for traversal, and practicality. The current Blackboard Architecture model does not organize facts. The only structure that the network has is supplied by the connections created by rules. From this network design, related facts on a large scale cannot be viewed as effectively by an end user, and that limits the practicality of the network.

Containers provide a storage method for related facts, which allows all facts within any container to be viewed, accessed, and manipulated at the same time. This previously nonexistent organization also provides a new method for traversal between facts, which has the potential for systems to traverse over fewer nodes, thus reducing traversal time.

This paper introduces and evaluates the addition of the concepts of containers and links to the Blackboard Architecture. Containers are used to group facts to by organizational, physical, spatial or other relationships. They are implemented as a new object type, which can be referenced and queried within the Blackboard Architecture. Links are used to model and describe the organizational, physical, spatial and other relationships between containers. Links and containers provide a way to group facts that is independent of the decision-making logic of the rule structure. They also facilitate the implementation of multiple levels of organization, as a fact can potentially be a member of multiple containers and link-container structures. Containers can also be nested to represent complex organizational, physical, spatial and other relationships, as needed. This additional organization and logic can be interrelated with the existing rule-fact networks, facilitating the accurate modeling of intricate, complex and understandable application-tailored. Both containers and links are described and evaluated herein.

This paper describes what containers and links contribute to the current Blackboard Architecture model and evaluates the impact of containers and links on network traversal time in varying scenarios. It shows that links and containers notably reduce the traversal time between facts in Blackboard Architecture networks in many circumstances.

This paper continues, in Section 2, with a review of several areas of prior work that provide a foundation for the current work presented herein. Then, Section 3 describes the implementation of links and containers. Section 4 provides a qualitative assessment of their efficacy. Following this, Section 5 presents the methodology used for performance analysis of links and containers and Section 6 presents the results of this analysis. Finally, the paper concludes, in Section 7, and describes potential areas of future work.

**2. Background**

This section reviews several areas of prior work which are relevant to this paper. First, the prior work on the Blackboard Architecture is reviewed. Then, prior work related to encapsulation using the Blackboard Architecture is discussed.

## 2.1. Blackboard Architecture

Hayes-Roth [1] proposed the Blackboard Architecture, in 1985, for use in decision making; however, it drew on prior work related to a speech recognition system [2]. The Blackboard Architecture builds upon the rule-fact network structures of expert systems [3], which were introduced in the 1960s and 1970s with the Dendral [4] and Mycin [5] systems. The initial system [1] had two separate components: one was designed to make use application domain decisions while the other made control-related decisions. Thus, the Blackboard Architecture adds the concept of actions [6] (i.e., the control decision-making), which can be used to interact with the system's operating environment, to the functionality commonly provided by expert systems.

The Blackboard Architecture has been used for numerous applications. Examples include robotics [7,8], detecting unmanned systems' failures [9], body area network communications [10], cybersecurity attack modeling [11], medical image interpretation [12], modeling proteins [13], tutoring [14], software testing [15], homeland security [16], making legal decisions [17], recognizing handwriting [18], planetary exploration [19], sound identification [20], making decisions with uncertainty [21], mathematical proofs creation [22], and poetry generation [23].

## 2.2. Encapsulation using the Blackboard Architecture

A variety of enhancements have been proposed to the Blackboard Architecture. Examples include the incorporation of pruning to expedite operations [24,25] and enhancements to support parallel [26] and distributed [27] processing. Augmented message filtering and processing [28] capabilities have also been proposed. Prior work also introduced an ability to solve blackboard [29,30] to facilitate goal-driven Blackboard Architecture uses.

One key area of prior work, which is built upon herein, was the use of boundary nodes [31] to encapsulate data from a region of a Blackboard Architecture network. This was designed to facilitate communications by increasing the value of the data being transmitted (e.g., by transmitting a conclusion or derived fact instead of simply raw data). Its use, in this way, was proposed for use in electronic warfare systems [32], body area network communications [10] and robotics [31].

This work built upon significant prior work related to multi-agent and distributed Blackboard Architecture implementations. The most basic of these implementations, proposed by Compatangelo [33,34], utilized shared memory as a form of communications, while allowing processing to be distributed between multiple CPUs on the same computer. Kerminen and Jokinen [35] proposed a more capable approach, which was able to command multiple systems; however, this implementation was limited by the use of a single storage mechanism. Redondo and Ortega [36] overcame the single storage location limitation through the use of a data checkout mechanism; however, this approach could be limited by concurrency and bandwidth limitations. Several hierarchical implementations [37–39] and a hub-and-spokes design [40–42] proposed solutions to aid with these concurrency and data transfer limitations. Approaches based on data distribution [43], replication [44], selective synchronization [45] and a client-server model [46,47] have also been proposed.

The "ambassador" data elements, described in [37,48] are of particular note. These elements were based in a hierarchical model and were representatives of subordinate areas in the blackboard of each superior level. This approach, though, is limited by the need to have a hierarchy and has the added complexity of using ad hoc direct messages to facilitate communications outside of the hierarchy.

The considerable history of distributed Blackboard Architecture implementations provides a foundation for the current work. The "ambassador" [37,48] and boundary nodes [31] are, perhaps, the most directly relevant. However, the current work inherently adds a new dimension to the Blackboard Architecture which goes beyond these concepts to support the implementation of a grouping mechanism for facts related to a single object (or concept) and a mechanism for representing physical (or other organizational) associations between the objects. This is inherently separate from – and potentially complementary to – solutions for distributed command, data replication and data concurrency management.

## *3. Links and Containers*

This section describes the two additions to the Blackboard Architecture which are introduced herein. First, in Section 3.1, containers are introduced and discussed. Then, in Section 3.2, links are presented.

### *3.1. Containers*

With the growing complexity of networks that are required to accurately represent modern-day systems, current Blackboard Architecture models struggle to provide understandable views of the data within these networks. Containers can help to improve both network searches and the understandability of the network by grouping related facts.

A container is an object, with the Blackboard Architecture, which has a description and a set of facts associated with it. The description of the container defines the relationship between the facts it references. This may include the name of the entity that it represents, or any defining term for the connection between its facts. Depending on the requirements of the system and the purpose of the container, different attributes can be stored in the containers. If the description field is not enough to uniquely define a container, an ID field can also be added. The container description, along with additional attributes, can be used during traversal, if desired, which can add a layer that has the potential to improve performance.

Facts can only be part of a single container. The purpose of containers is to create a hierarchical storage structure that can be layered on top of an existing blackboard network, rather than an additional type of connection. To support a container's ability to create clean, compartmentalized views of facts, they require clear boundaries between the facts that are in the container and those that are not. Facts are grouped by the container that they are a part of and are not all independently connected, as with rules. Connections between containers, however, are possible.

These connections are created using links, which are described in the next section.

**3.2. Links**

While containers are used to organize facts, links are objects within the Blackboard Architecture which are used to create connections between these containers. With these connections, groups of facts can be modeled as being related to each other, thus providing more organizational complexity to the network, without the significant processing overhead and implementation awkwardness that would be incurred by trying to define these relationships using only unorganized collections of facts and rules.

Each link stores a description. Similar to containers, the description indicates the relationship between the entities it connects (in this case, the relationship between the two containers). In addition to the description, two references to containers are stored. These are the two containers that are linked by the relationship identified in the description of the link. One of the containers referenced is identified as the start container, and the other container is the end container. As these names suggest, links have specified directions. These directional links can be used for many purposes, ranging from a representation of a logical traversal to representing a hierarchical parent-child relationship. Bidirectional relationships can be readily defined using two individual links between the two containers, with the containers acting as the start and end containers switched between the two links.

Directionality in links is necessary within precisely configured networks due to the implications of bidirectional links. Bidirectional links would imply that a container can be traversed in both directions, which is often not the case. As rules have input facts and output facts that indicate a direction for traversal, unidirectional links provide an explicit direction that must be followed during traversal. If links were bidirectional, that would result in potentially unwanted traversals being completed if the relationship was actually only unidirectional. In hierarchical situations, one link may be created to indicated that container 1 is a parent of container 2, and another link could be used to indicate that container 2 is a child of container 1. If links were bidirectional, there would not be any indication of which container was the parent and which was the child. Relationships are often one-directional in this sense.

**4. Assessment of the Efficacy and Benefits of Using Containers and Links**

This section evaluates the efficacy and considers the benefits and drawbacks of using containers and links.  First, they are compared to using only facts and rules.  Then, applying rules to containers is discussed.
Third, their utility for enhancing systems' organization and usability is discussed.  Following this, their role in network traversal is discussed.  Finally, their benefits to system implementation are evaluated.

*4.1. Contrast to Fact and Rules*

There are many parallels between containers and links and facts and rules. Containers may store attributes that could behave similar to facts, and as rules connect facts, links connect containers. It is in their functions that their differences and benefits are shown. Rules show the logic between facts; they dictate values. When input facts are a specific value, that results in an output fact's value changing. Links and containers do not, in themselves, hold any logic. They simply organize data, create relationships to be used in other network functions, and make the organization of the network easier. One use for containers and links is to show relationships between real-world entities. For example, a laptop may be connected to a router. There is no inherent logic between those two entities, so each would be represented by a container. A link could then connect them. Rules and facts would only play a part in this relationship if there were rules between the containers.

### 4.2. Applying Rules to Containers

Rules are used to represent a logical relationship between facts in a Blackboard Architecture network. The same logical may apply across containers. In the current Blackboard Architecture model, rules may include any facts that are available. However, rules work differently with containers. To implement a rule across two containers, the facts involved in the rule may only be from the two connected containers. When using binary facts, for example, a rules could be created to indicate that if fact 1 is true in container A and fact 2 is false is container B, then facts 3 and 4 will be true in container A, and fact 5 will be false in container B. Rules can still serve the same logical purpose that they normally do with facts; however, by restricting the scope to container connections, additional capabilities are added. If a rule is triggered due to its input fact values being changed, the containers involved in the rule are taken to have their facts changed, rather than just disassociated facts having their values changed. In the previously mentioned example of a laptop being connected to a router, a rule could be developed to indicate that if the laptop has the password to the router, then the router will provide internet connectivity for the laptop.

### 4.3. Organization and Usability

The main purpose of links and containers is to provide an additional layer of organization within a Blackboard Architecture network. A network that is comprised of only facts and rules stores its knowledge in facts and relies upon the rules to provide both structure and functionality. In many cases, the rules that are needed for providing system functionality do not represent the logical connections between groups of facts in a meaningful way. Instead, they simply provide functions that dictate the values of required input and targeted output facts.

The addition of containers facilitates having an organizational structure that is independent of operational rules. By separating facts into containers, they can be organized in a way that makes sense to users. They can also be separated based upon shared characteristics, geographic location or by any other application-relevant organizational system. Notably, applications may make use of these containers and the fact values within them in a variety of ways. They could be used for localized or comparative decision-making. Alternately, custom code could be developed, for an application, to allow fact values within a region to be set and updated as a group or for values to be set, updated or queried selectively within groups. Notably, a key function of Blackboard Architecture networks is the ability to readily get information from the network. By implementing containers, the network itself provides a way to group facts, which can improve the understandability, usability and performance of the system, as well as making it suitable for certain applications that the base system may not be readily useful for.

Containers, thus, have the ability to provide more transparency within a Blackboard Architecture network. Facts' values can be read within the typical Blackboard Architecture network. However, their organization (or perhaps, more correctly, lack of explicit organization) may obfuscate or deter operators from seeing relationships between associated facts or facts which have corresponding roles with regards to similar objects. Using containers, operators can quickly access related facts and compare corresponding ones. This facilitates both human and automated assessment to identify patterns and potential implementation errors. This additional layer of coherence can make a notable difference when working with larger networks, where operators may be trying to find meaning from thousands or even millions of fact data points.

Links serve a similar organizational purpose. They show the relationship between container groups in a way that rules, which are designed to have an operational interaction with facts, cannot. With the ability to store a name or description in a link, connections between separate groups of facts can have a defined meaning. When viewing, assessing or performing operations on a Blackboard Architecture network, links allow containers connections to be readily apparent and implemented in a manner that is understandable to both humans and software. Links can also aid in path-finding through the organizational network, in addition to or instead of solving Blackboard Architecture networks for goal completion pathways.

*4.4. Traversal*

Links create connections between containers, which allows them to be used as a means for exploration, analysis and traversal of the network. With the categorization of facts into individual containers, a new mechanism is created to reference those facts. Instead of searching simply for an individual fact within a network, a container can first be searched for. This has the potential to significantly reduce the search time within the system as well as traversal path calculation time from one fact to another. Applications can also use relevant containers and links as part of the operational logic. For example, a system might need a particular capability (i.e., a particular fact asserted within a container); however, the exact provider of the capability may not matter. Any container with a corresponding asserted fact, thus, could facilitate system operations, as opposed to only a single fact being given this ability, within a rule-fact network, or numerous rules having to be developed to allow any one of numerous facts to satisfy the criteria.

From a traversal perspective, the process can be simplified significantly, in many cases, using links and containers. Presuming that most containers hold several facts, the containers can become a primary traversal mechanism, rather than the system having to traverse through every possible fact between the start and end facts. If an application area makes use of an organizational system implemented with containers, a start fact can reference its parent container, and links can be used to traverse through the network to the container that holds the end fact. Presuming that containers each contain multiple facts, container traversals inherently will take less computational time, simply because there are less nodes and less interconnectivity complexity between the start and end facts.

The traversal benefits of links and containers, of course, depend on the configuration of the Blackboard Architecture network they are being used with. A traversal can only be done between two facts via containers if the facts are placed in containers and those containers are interconnected with links. The number of facts in each container and the level of interconnectivity between containers, as opposed to what would be required for an implementation using only rules and facts, also drive the level of processing time reduction that will be enjoyed. Of course, numerous other application-specific considerations, including what rules, facts, containers and links represent in the real world and how they're used in the modeling system also drive the extent of the benefit that containers and links will provide, in any particular instance.

*4.5. Implementation*

Containers and links have practical uses in many application areas. Containers ability to hold several facts creates a layer of organization that can significantly increase the functionality of a Blackboard Architecture network. Two examples of how containers and links can be used in specific Blackboard Architecture networks are presented in this example. First, an example regarding employee data is presented. The, the use of links and containers for a Blackboard Architecture system modeling a computer network for analysis is discussed.

*4.5.1. Example – Employee Data*

At most companies, it is essential that employees' data is stored accurately, efficiently, and usably. This information may include the employee's name, job title, direct manager, and office location, among other attributes. Using a Blackboard Architecture network, facts can be used to represent each piece of data that an employee has. However, without links and containers, all the facts for any given user would not be related. A fact would need to state, in its entirety, for example, "John Doe works in the Sacramento office". This would need to be done for every employee, and those working out of the Sacramento office would not be known to the system to be related.

Containers and links can provide much-needed organization to this situation. Separate containers can be created for each user and each office location. Each user can have their own facts that represent their characteristics rather than having all the facts as one big unorganized group. They will each be within their own compartmentalized containers. The Sacramento office can have its own container, storing details such as its official name and address. Then, the container representing John Doe can be linked to the Sacramento office, via a link, with a description of "works in". By creating these containers and links for all Sacramento office employees, they can all be related via the common Sacramento office container.

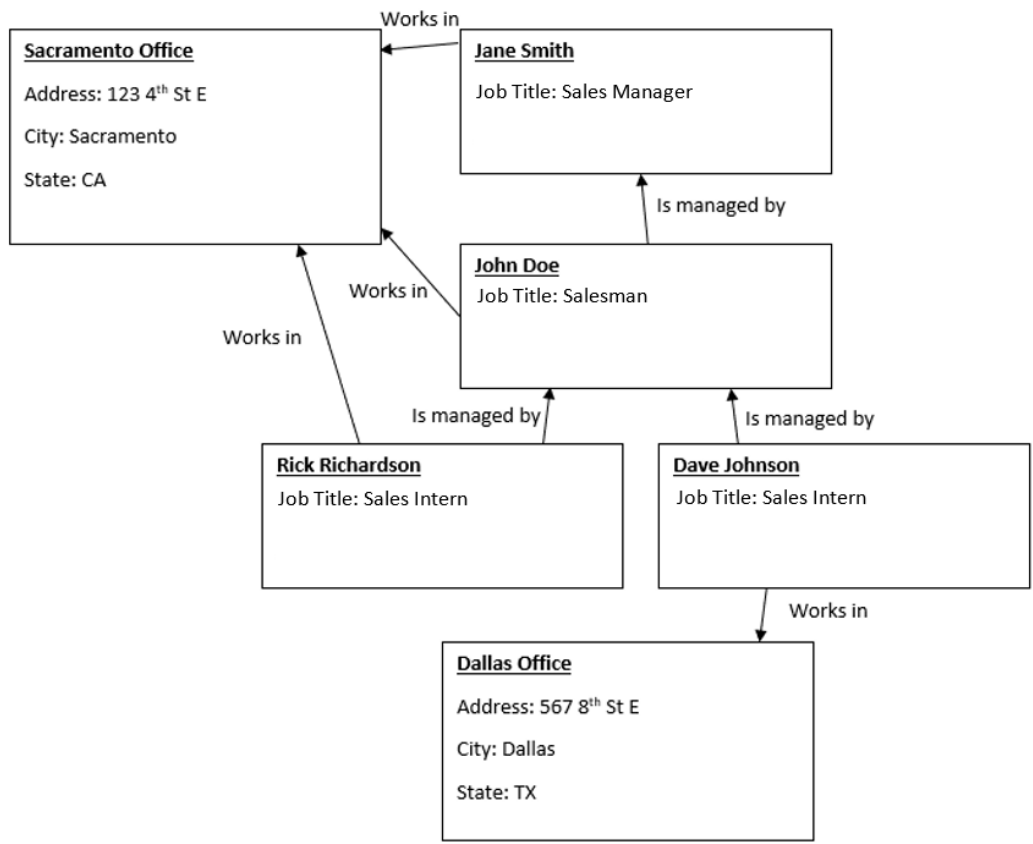

*Figure 1: Employee Data Example Blackboard Network*

In addition to the office location link, a link can be created between John Doe and his direct manager. This would allow the relationship between John and his manager to be stored and traversed. However, the unidirectionality of the link does not have to restrict traversal. Even if the link represents that John's manager "manages" John, the link could also be used to search from John back to his manager. The function of links during traversal is not limited by the data stored in the link. A query of all of John's incoming links that indicate management would return every employee that is considered a manager of John. The current Blackboard Architecture model would not provide nearly as much relational organization between employees. Without the use of containers to organize facts and links to connect the containers, facts would need to be created to match every user with their manager. When a manager was to be queried using the name of the employee, every rule would need to be searched. With containers, the container for a user with the specified name would need to be found, and the manager could easily be determined by searching through the incoming links.

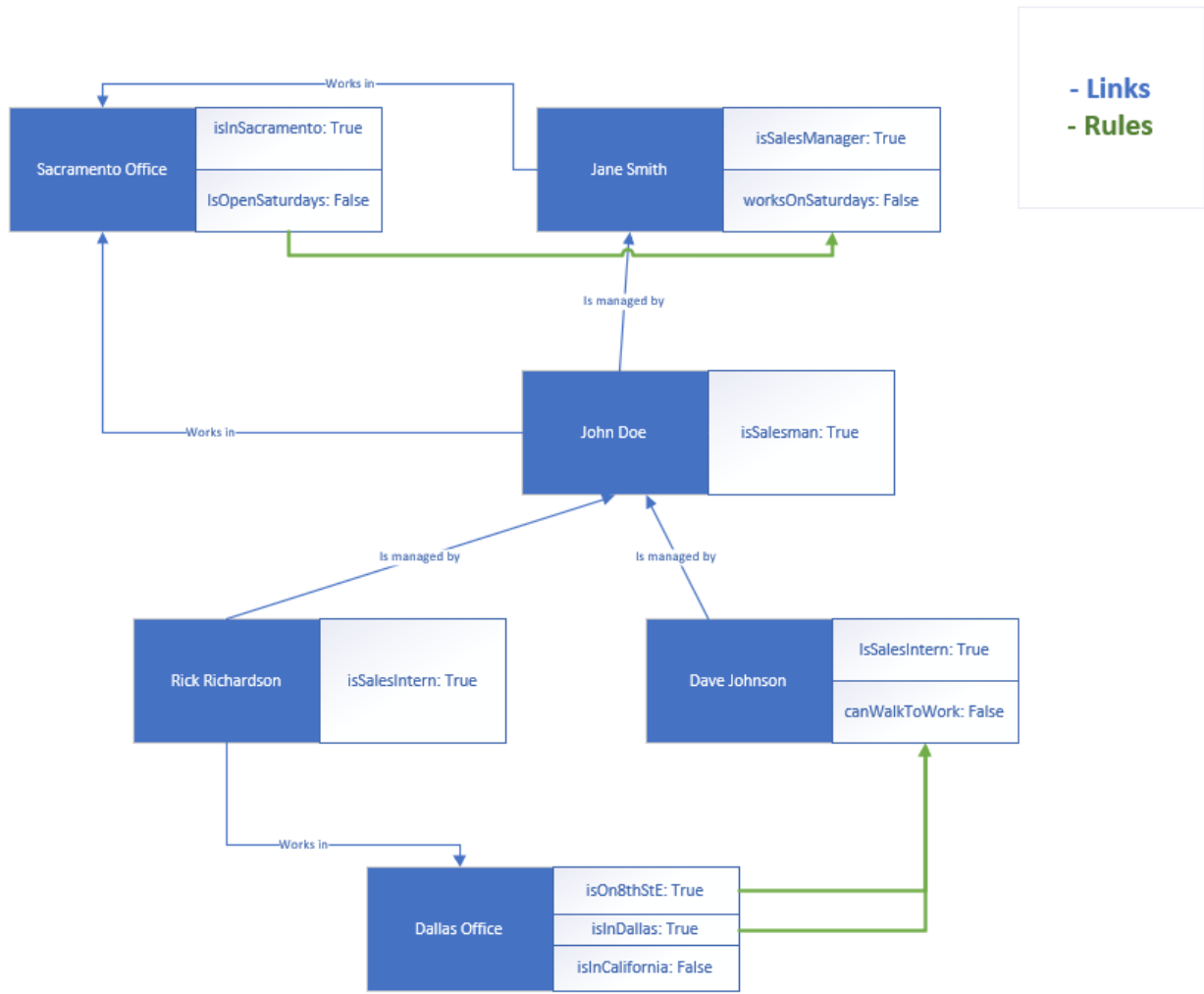

*Figure 2: Object Diagram of Rules, Facts, Links, and Containers for Employee Example*

The implementation of this in a Blackboard Architecture network is shown in Figure 2.

The object diagram shows the way that containers, links, rules, and facts could interact in this example. The shaded boxes show the description of each container, while the sectioned boxes, to the right, show the fact descriptions along with their corresponding values. The green arrows represent rules between the facts, and the blue arrows represent links, with their descriptions labeled. By combining all the data from these objects, one can learn a large amount of information from the network.

In the current model, facts are binary objects, which store values of true or false. Because of this, any non-binary value from a real-world scenario must be converted into an appropriate binary format. In some cases, such as with addresses, it is the relevant data from the address that is extracted and used for facts. The reason that all the information may not be needed for a rule is because that rule's value only depends on a binary value that can be extracted from the real-world value. For example, a rule determines whether Dave Johnson can walk to work. In this example, if Dave lives on 8$^{th}$ Street in Dallas, business logic may indicate that he may only walk to work if its address is also on 8$^{th}$ Street in Dallas. Therefore, those are the parts of the address that are included as facts, and those facts will determine whether or not Dave walks to work.

In addition to this rule, another rule can be created regarding the Sacramento office being open on Saturday and Jane Smith working on Saturday. Since the preconditions of a rule can affect the postconditions for that rule, if the Sacramento office were to start being open on Saturday, Jane would then start working on Saturdays. By combining the rules, links, containers, and facts, one could determine that "Jane Smith does not currently work on Saturdays at the Sacramento office. If the office were to open on Saturdays, though, she would start working on Saturdays." This requires a quick traversal from a start container (Jane Smith) over the link described as "works at" to the Sacramento office container, showing that Jane works at the Sacramento office. Then, by using the rule, the latter part of the sentence can be concluded: that Jane would start working if the office was open on Saturdays.

*4.5.2. Example – Network Information*

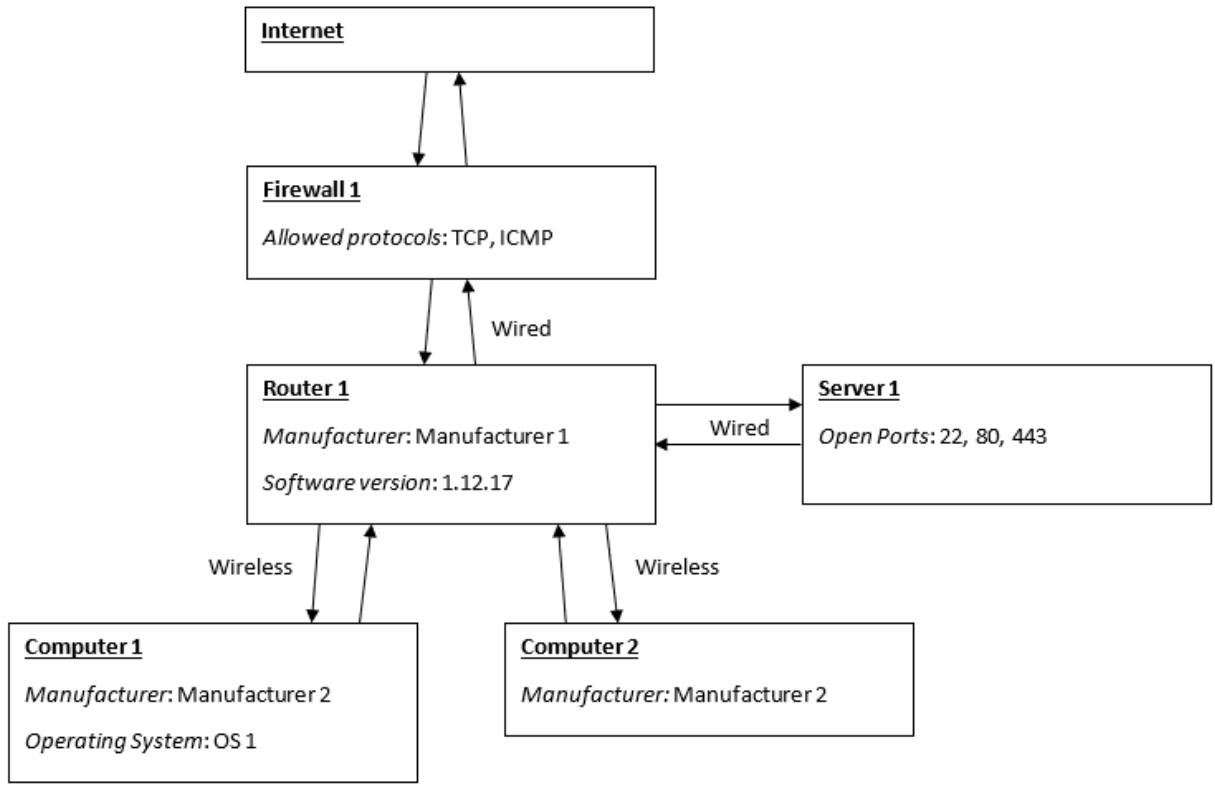

*Figure 3: Networking Equipment Example Blackboard Network*

The connections between network equipment can also be modeled by containers and links. In this example, a firewall is connected to a router, which is connected to a server via a wired connection and two laptops wirelessly. These devices are represented as containers, each containing their own facts. In a network, different information is often known about each device. In Figure 3, it is shown that both computer 1 and computer 2 have the same manufacturer (manufacturer 1), but only the operating system on computer 1 is known. By having a container for every device, facts can be organized in a way that was not possible with previous Blackboard Architecture network versions. If the network was simply constructed with facts and rules, every fact would need to contain information about the device it belongs to in it, resulting in long names such as "Router 1 Manufacturer" or "Firewall 1 Allowed Protocols" in order to associate data with particular devices.

It is also important to note that containers do not necessarily need to hold facts. In the example in Figure 3, it is important to know that Firewall 1 has access to the internet, but it is not necessary to store any additional information about the internet as the connection itself is enough information to provide the needed information.

Links can be used in networks to represent data interconnections between different devices. Links have descriptions that can be used to provide information about the link. In this example, the descriptions state whether the link is a wired or wireless connection. This provides additional information about the network that can be used in traversal.

Traversal of this network could be used in many ways. If someone wanted to know if computer 1 could retrieve a web page from server 1, they could determine that there are links in both directions starting at computer 1, going through router 1, and reaching server 1. This indicates that a request could be sent to server 1, and the web page could be returned back to computer 1. However, if a UDP packet was to be sent from computer 2 through firewall 1, that packet would not make it through. This can be determined by comparing the packet's protocol against the allowed protocols for firewall 1. This shows that UDP is not listed. Therefore, the packet would not be allowed through, and one can conclude that the movement of that packet through the network would end there.

The fidelity of networks created via containers and links is limited only by the data available, so if more information became available about the devices in Figure 3, that information could be added to the containers in the form of additional facts. If network traversability depends on the fact values of different containers, the addition of facts and adoption of rules to reflect this could affect traversal through the network, thus providing a more robust view of the system represented by the network.

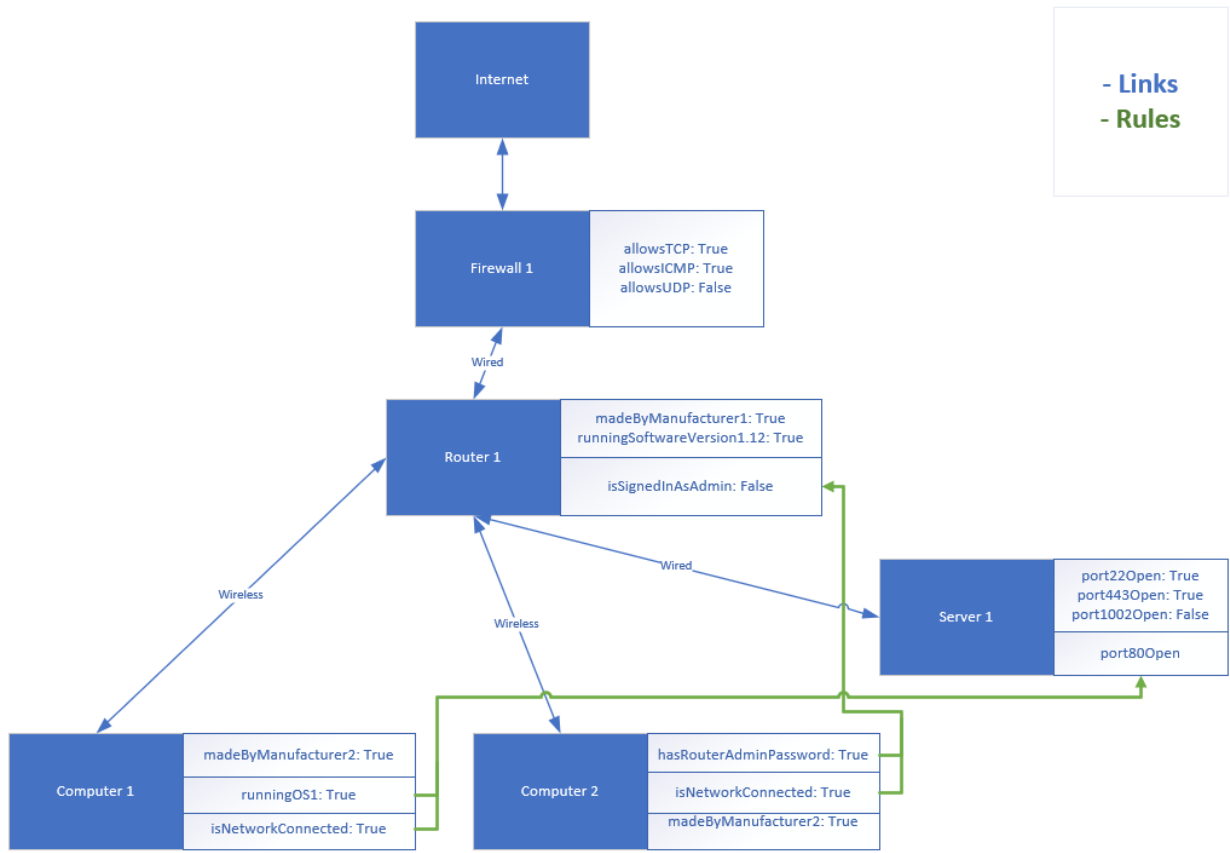

*Figure 4: Object Diagram of Rules, Facts, Links, and Containers for Networking Equipment*

Figure 4 shows the Blackboard Architecture implementation of the network shown in Figure 3.

As shown, rules can be created between the same containers that links exist between. For example, router 1's fact indicating whether an administrator is signed in is determined by computer 2. If that computer is network connected and has the router's password, then the router can be signed into as an admin, thus changing that fact value to true.

It is relationships like these that facilitate modeling, in an understandable way, growing complexity within Blackboard Architecture networks. Not only can the properties of the containers be considered during traversals, but rules can be used too. By checking if a specific rule exists between the facts in two adjacent containers, the evaluation of that rule could be used to determine whether a link should be traversed or not. The usability of rules depends on the application that the network is being used for, as different logic may be implemented for different scenarios.

**5. Performance Testing**

This section discusses the performance testing that was performed to assess the use of containers and links. First, an overview of the testing system is provided. Then, the testing variables of container assignment method (Section 5.2) and network scaling ratio (Section 5.3) are discussed. Finally, Section 5.4. discusses the performance testing process that was used.

*5.1. Testing System Overview*

To demonstrate the effectiveness of integrating containers and links into the current Blackboard Architecture framework, a system was constructed that implemented these concepts to facilitate data collection for analysis. The system uses Blackboard Architecture networks to compare traversal, in terms of both the time required and the number of node traversals performed. Traversal using rules is compared to traversal using containers and links. The performance of the two approaches was compared under different experimental configurations. The base configuration, which other configurations are created from by varying parameters, uses the below configuration options:

Facts: 1000     Rules: 1000     Containers: 100     Links: 400

*5.2. Container Assignment Methods*

Testing was performed to analyze the performance impact of the distribution of facts stored in containers. Three different container assignment methods were tested which are now discussed.

The first method, uniform assignment, evenly assigns facts to each container. This is done by looping through all the containers in the network and assigning one randomly selected fact at a time until all facts have been assigned. This results in containers that have a maximum difference in the number of facts per container of one.

The second assignment method is random assignment. In this method, each fact is assigned to a randomly selected container. This potentially allows any portion of the facts to be in any of the containers. At the most extreme, some containers may have many facts, while other may have none assigned to them.

The final assignment method is loaded assignment. For this method's assignment, a random number between 1 and the number of containers in the network (inclusive) is generated. That random number is then used to select that number of containers and uniformly assign all the facts in the network to them. This creates a "loaded" network that has all the facts in a portion of the containers and no facts in the remainder of the containers. These assignment methods facilitate the analysis of the comparative operations of the Blackboard Architecture under multiple configurations which may be relevant to different real world application area needs.

*5.3. Network Scaling Ratios*

To test the impact of links and containers on networks of varying sizes, the network was scaled by applying several scaling factors to the base network configuration (presented in Section 5.1). The following scaling factors were utilized:

50%, 75%, 100%, 125%, 150%, 200%

For each scaling factor, the number of units of each entity (facts, rules, containers, and links) was changed, according to the given percentage of its original value. For example, the test evaluating the influence of fact count on the traversal speed of the network would start with 50% of the control configuration for facts (1000), which would result in 500 facts being used in the network. Once all iterations were completed for that test, 750 facts could be used in the network, and so on. Once the rest of the entities (rules, containers, and links) were scaled at 50% individually in the same fashion, all entities were scaled at 50% simultaneously. This resulted in a network containing 500 facts, 500 rules, 50 containers, and 200 links.

*5.4. Testing Process*

This section describes the testing process that was used. First, the tests that were used are described in Section 5.4.1. Then, a single testing run is defined in Section 5.4.2.

*5.4.1. Tests*

For each container assignment method, tests were run using each scaling percentage. First, each network parameter (facts, rules, containers, and links) was scaled individually, according to the scaling percentage, while the other parameters remained at the control number. Once the individual scaling of each parameter was completed, concurrent scaling of all parameters was performed, thus evenly scaling the entire network. For each experimental condition, 100 tests were performed.

*5.4.2. A Single Testing Run*

For each test run, a randomly generated Blackboard Architecture network was created. This network contained facts, rules, containers, and links. Two random facts, which could be traversed by both rules and by links and containers, were selected. A traversal was performed between the two facts, first, via rules, and then, by using links and containers. The time elapsed (in ticks) during the traversal was collected and compared.

The results from the experimentation described in this section are presented in Section 6.

**6. Data and Analysis**

A total of 90 different network configurations were tested, with a total of 9000 tests performed. One traversal via rules and another via links was performed for each test resulting in 18000 traversals being completed. Through the testing, it was found that, as expected, traversing a network via links and containers is faster, on average, than simply traversing via rules.

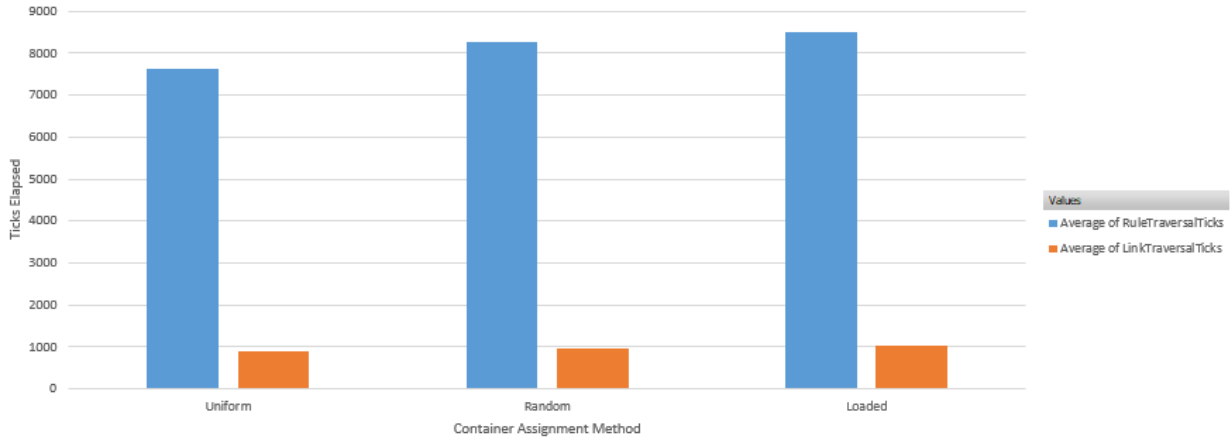

*Figure 5. Container Assignment Method vs. Ticks Elapsed (Original Scale).*

*Table 1. Data for Container Assignment Method vs. Ticks Elapsed (Original Scale)*

| Container Assignment Method | Average of Rule Traversal Ticks | Average of Link Traversal Ticks |
|---|---:|---:|
| **Uniform** | 7622.248 | 904.13 |
| **Random** | 8265.048 | 971.71 |
| **Loaded** | 8494.558 | 1018.284 |

Figure 5 and Table 1 show the average amount of time, in ticks, that a traversal took between two randomly chosen start and end containers. The average traversal time for a rule-based traversal with uniformly-distributed facts was 7622.248, whereas link-based traversals averaged 904.13 ticks.

This relationship was consistent for every container assignment method. It consistently took approximately eight times the amount of time to traverse via rules than to traverse via links and containers, at the original (100%) scale. Figure 5 also shows a slight increase across the container assignment methods, from uniform to random to loaded. Since the containers and rules do not interact, the container assignment method should have no direct impact on the traversal times via rules. However, the link traversal time is impacted by the container assignment method. Each container assignment method impacts the distribution of the network's facts within the containers. This makes it either more or less likely that the two selected facts will be in the same or more directly connected containers. For example, for a uniform distribution, each container has the same probability of containing one of the facts. However, for the loaded distribution, only a portion of the containers may contain one of the facts. Therefore, the loaded container assignment method has a greater chance of having fewer links that connect the containers of the start and end facts.

Figure 5 and Table 1 use the control values for each of the network entities. Figures 6 and 7 and Tables 2 and 3 show the performance of the network at 50 and 200 percent of the control entity size.

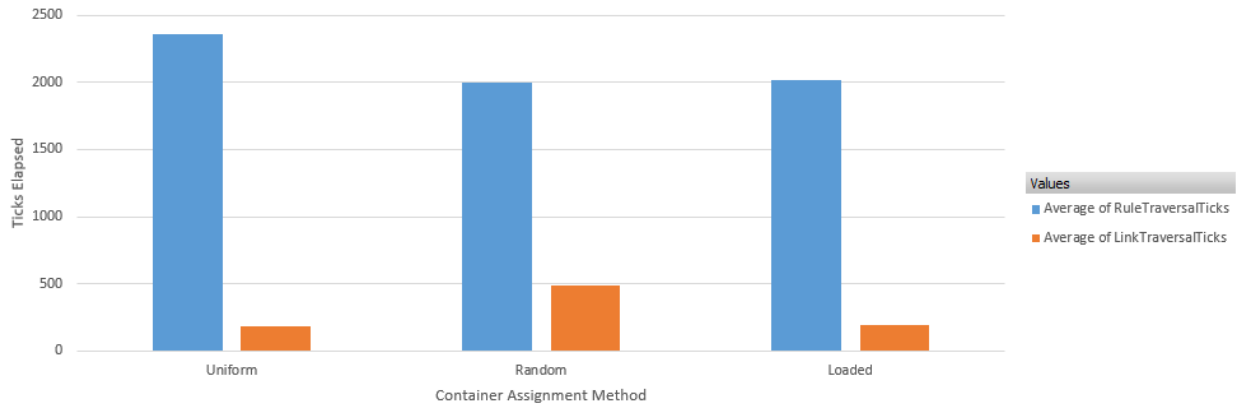

*Figure 6. Container Assignment Method vs. Ticks Elapsed (50% scale)*

*Table 2. Data for Container Assignment Method vs Ticks Elapsed (50% scale)*

| Container Assignment Method | Average of Rule Traversal Ticks | Average of Link Traversal Ticks |
| --- | --- | --- |
| **Uniform** | 2358.07 | 182.45 |
| **Random** | 2000.54 | 489.59 |
| **Loaded** | 2020.4 | 191.76 |

At 50% of the original size, the traversals, as would be expected, take less time. With a smaller network size, it can be seen that the ratios are more sporadic. At the original scale, the ratio of time spent traversing via rules versus links was consistently about 8:1. However, with a 50% scale, the ratios range from roughly 4:1 to 10:1. Figure 6 shows that link traversals still take less time that rule traversals.

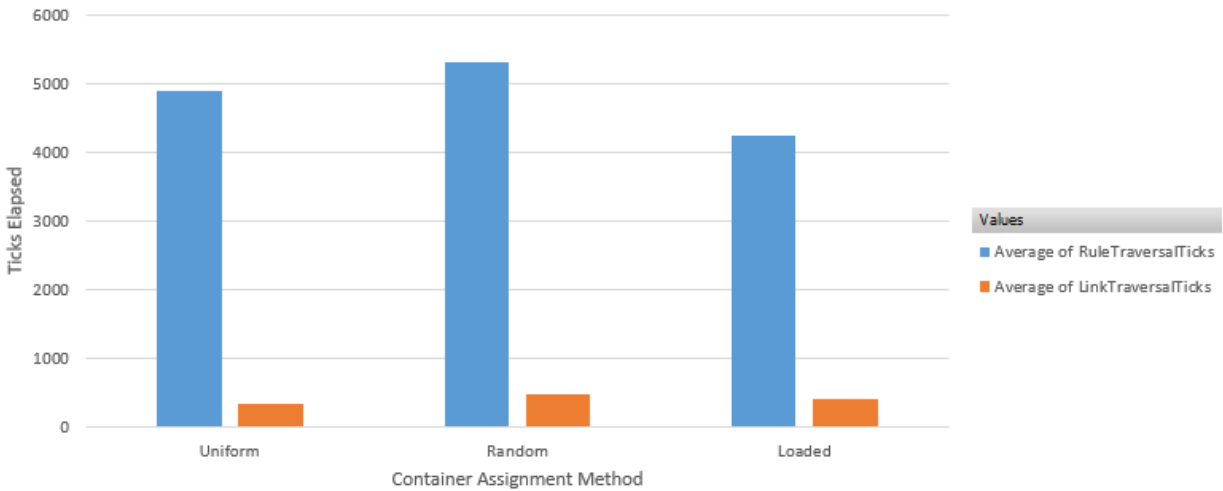

*Figure 7. Container Assignment Method vs. Ticks Elapsed (75% scale)*

*Table 3. Data for Container Assignment Method vs. Ticks Elapsed (75% scale)*

| Container Assignment Method | Average of Rule Traversal Ticks | Average of Link Traversal Ticks |
|---|---|---|
| **Uniform** | 4887.29 | 345.64 |
| **Random** | 5301.48 | 468.79 |
| **Loaded** | 4232.65 | 400.78 |

It is shown in Figure 7 that, in relation to the average rule traversal times, the average link traversal time has decreased. The random traversal over links has even decreased overall, from an average of 489.59 ticks to an average of 468.79 ticks. This shows that the random container assignment does have an impact on traversal, as there is a higher possibility that there is an irregular distribution for this approach, as compared to the uniform distribution approach. The uniform and loaded distributions were more consistent, with the uniform assignment time required staying slightly under the loaded time required, as was also the case at the 50% scale.

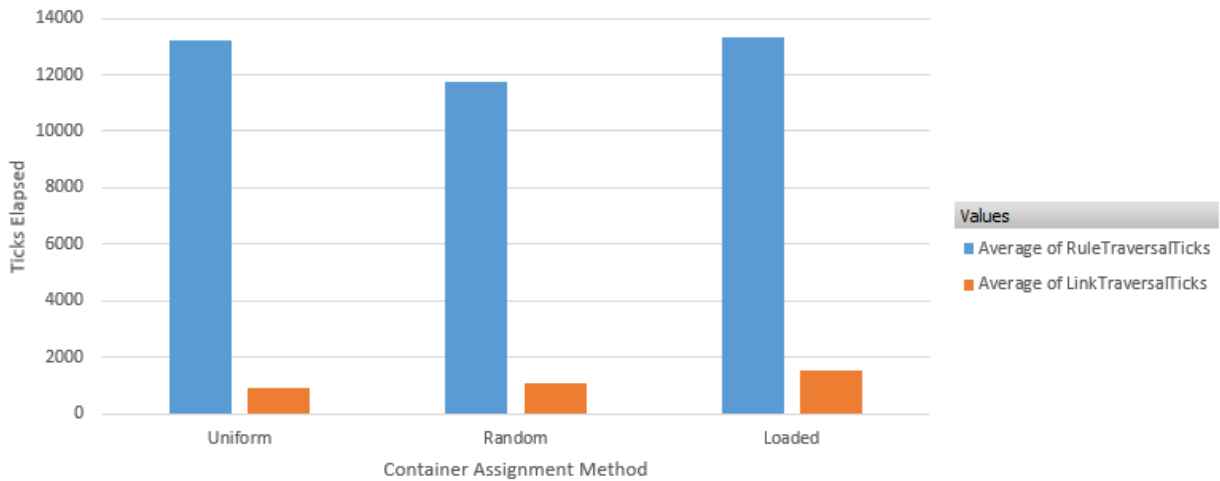

*Figure 8. Container Assignment Method vs. Ticks Elapsed (125% scale)*

*Table 4. Data for Container Assignment Method vs. Ticks Elapsed (125% scale)*

| Container Assignment Method | Average of Rule Traversal Ticks | Average of Link Traversal Ticks |
|---|---|---|
| **Uniform** | 13214.95 | 896.22 |
| **Random** | 11728.62 | 1090.96 |
| **Loaded** | 13302.54 | 1537.54 |

With the network scaled above the 100% values, Figure 9 and Table 5 show that the uniform and loaded rule traversal time requirements are very close to each other. However, there is a much large discrepancy between the link traversal time between the two distributions. In addition to this, both the uniform and random container assignment methods have link traversal times that are less than one-tenth of their rule traversal times.

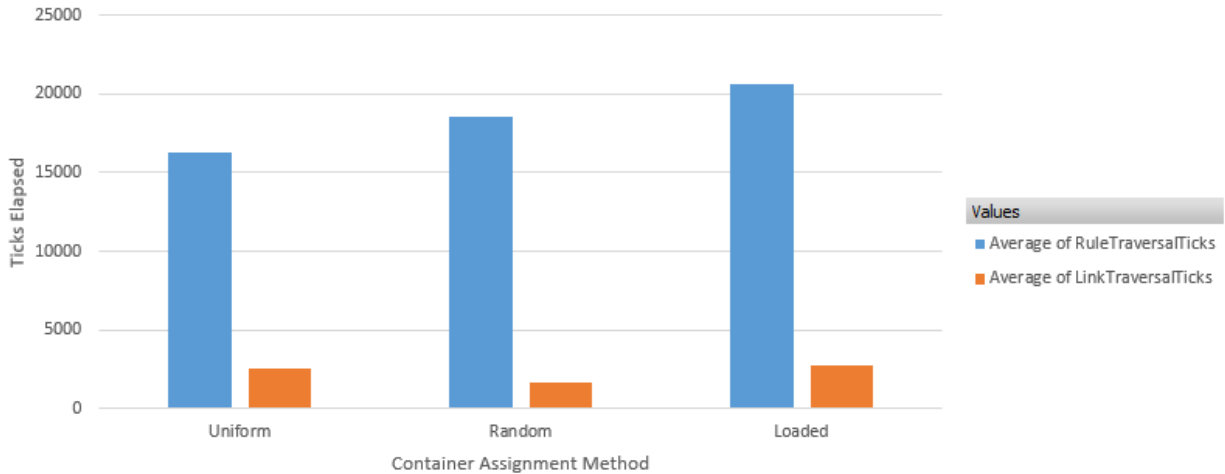

*Figure 9. Container Assignment Method vs. Ticks Elapsed (150% scale)*

*Table 5. Data for Container Assignment Method vs. Ticks Elapsed (150% scale)*

| Container Assignment Method | Average of Rule Traversal Ticks | Average of Link Traversal Ticks |
|---|---:|---:|
| **Uniform** | 16304.06 | 2522.73 |
| **Random** | 18522.98 | 1628.69 |
| **Loaded** | 20639.89 | 2776.69 |

It is shown in Figure 9 and Table 5 that for the 150% scale, the differences between the rule and link traversal times were still present but remained approximately the same overall, between all container assignment methods. The difference decreased proportionally for the uniform distribution and the loaded distribution, but decreased slightly for the random distribution, as compared to Figure 8.

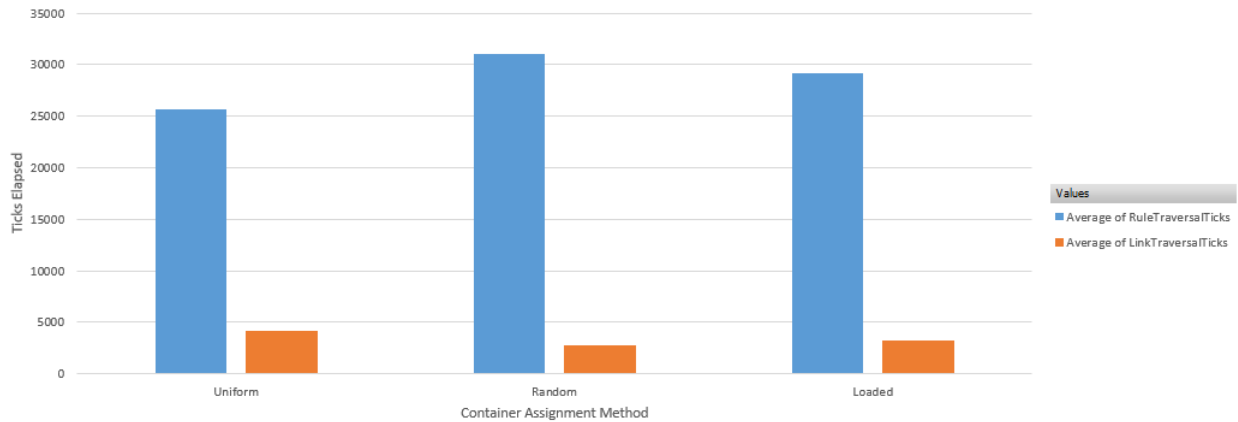

*Figure 10. Container Assignment Method vs. Ticks Elapsed (200% scale)*

*Table 6. Data for Container Assignment Method vs. Ticks Elapsed (200% scale)*

| Container Assignment Method | Average of Rule Traversal Ticks | Average of Link Traversal Ticks |
|---|---:|---:|
| **Uniform** | 25711.15 | 4183.74 |
| **Random** | 31014.28 | 2786.62 |
| **Loaded** | 29201.73 | 3284.71 |

As with the other two scaled networks, Figure 10 shows that, at a scaling factor of 200%, the link traversal still took less time than the rule traversal, for the random container assignment method. The link traversal time was also shown to be smaller, as compared to the rule traversal time, relative to the other two scaled networks.

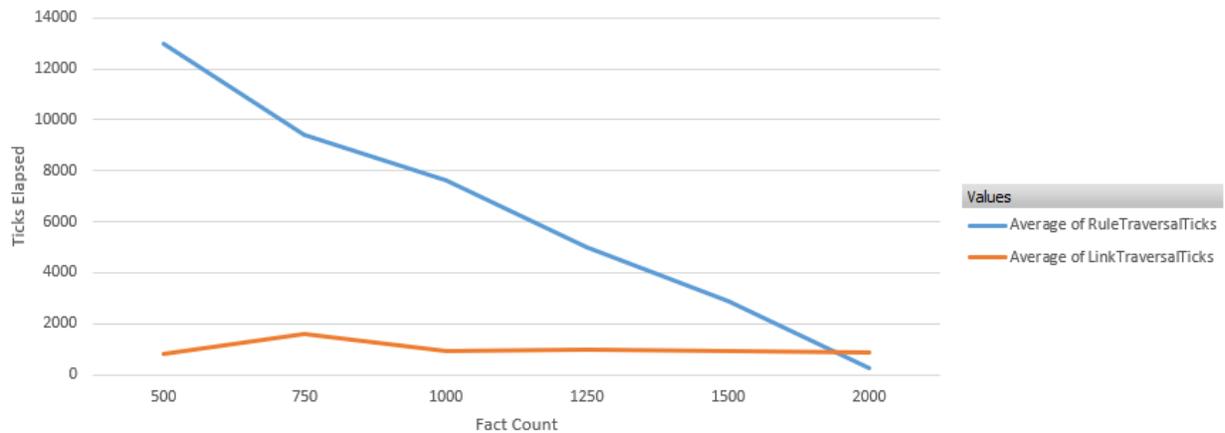

*Figure 11. Fact Count vs. Ticks Elapsed*

*Table 7. Data for Fact Count vs. Ticks Elapsed*

| Fact Count | Average of Rule Traversal Ticks | Average of Link Traversal Ticks |
|---|---:|---:|
| **500** | 12972.5 | 791.87 |
| **750** | 9389.45 | 1617.12 |
| **1000** | 7622.248 | 904.13 |
| **1250** | 4973.77 | 973.25 |
| **1500** | 2879.53 | 917.59 |
| **2000** | 263.43 | 882.11 |

Figure 11 shows the impact that adding facts has on rule and link traversal times in a network. In the tests, two random facts from the network, that can be traversed both by rules and links, are selected. The distribution of rules and links is random. Links connect containers, and containers only hold facts – they do not create traversable links between them. Because of this, the number of facts in the network does not influence the traversal speed via links across the network.

Rules, on the other hand, do create traversable connections between facts. It could be expected that a growing number of facts would increase the time required for rule-based traversal. This would be the case if the number of rules scaled with the number of facts, as shown in Figures 9 and 10. However, in this experiment, the number of rules stays consistent, which, with their random distribution, means that less facts will be directly connected. To perform a traversal (and be valid for the test), the two facts must be traversable via both links and rules. A larger number of facts, relative to rules, means that it is more likely that some facts will be left uninvolved in any rules, thus resulting in more disconnected strings of rules. It is more probable that facts are directly connected via a single rule in this case, which, as the fact count overtakes the rule count, results in lower rule traversal times for connected facts, but lower connectivity overall.

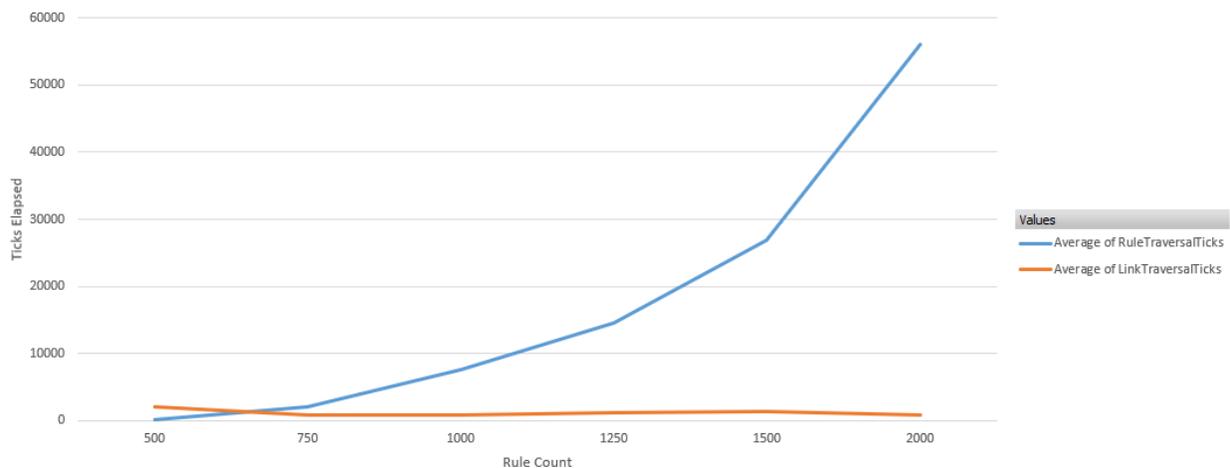

*Figure 12. Rule Count vs. Ticks Elapsed*

*Table 8. Rule Count vs. Ticks Elapsed*

| Rule Count | Average of Rule Traversal Ticks | Average of Link Traversal Ticks |
|---|---:|---:|
| **500** | 143.71 | 2099.96 |
| **750** | 2081.32 | 877.45 |
| **1000** | 7622.248 | 904.13 |
| **1250** | 14621.92 | 1168.64 |
| **1500** | 26917.51 | 1348.54 |
| **2000** | 56027.26 | 805.17 |

In contrast to Figure 11, Figure 13 shows what happens when more rules are added to a network. The ratio of rules to facts becomes higher, resulting in longer strings of rules connecting facts and requiring more time to search through all the rules. However, the number of links and containers stays the same, which is why the link-based traversal time remains relatively constant.

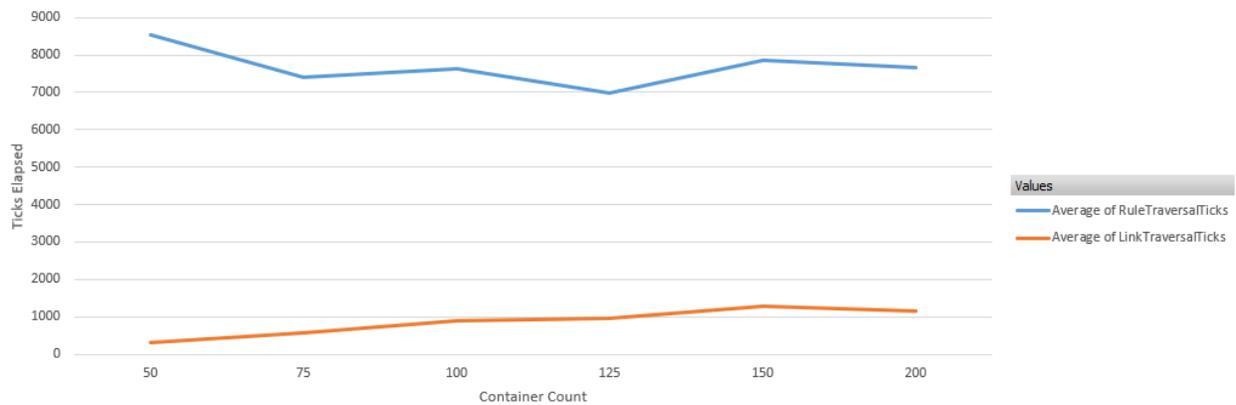

*Figure 13. Container Count vs. Ticks Elapsed*

*Table 9. Data for Container Count vs. Ticks Elapsed*

| Container Count | Average of Rule Traversal Ticks | Average of Link Traversal Ticks |
|---|---:|---:|
| **50** | 8525.35 | 310.93 |
| **75** | 7420.57 | 557.38 |
| **100** | 7622.248 | 904.13 |
| **125** | 6969.71 | 951.88 |
| **150** | 7866.62 | 1289.87 |
| **200** | 7650.34 | 1144.26 |

Figure 13 shows that the container count does not have a noticeable impact on the rule-based traversal time. This is because the two are largely unrelated. However, since links connect containers, adding containers is shown to slightly increase the link-based traversal time. This increase is likely due to the increased chance that the two selected facts are in different containers. With fewer containers, there are more facts in each container. When two facts are randomly selected, having less containers creates a greater chance that they are in the same container, which does not require a traversal over a link.

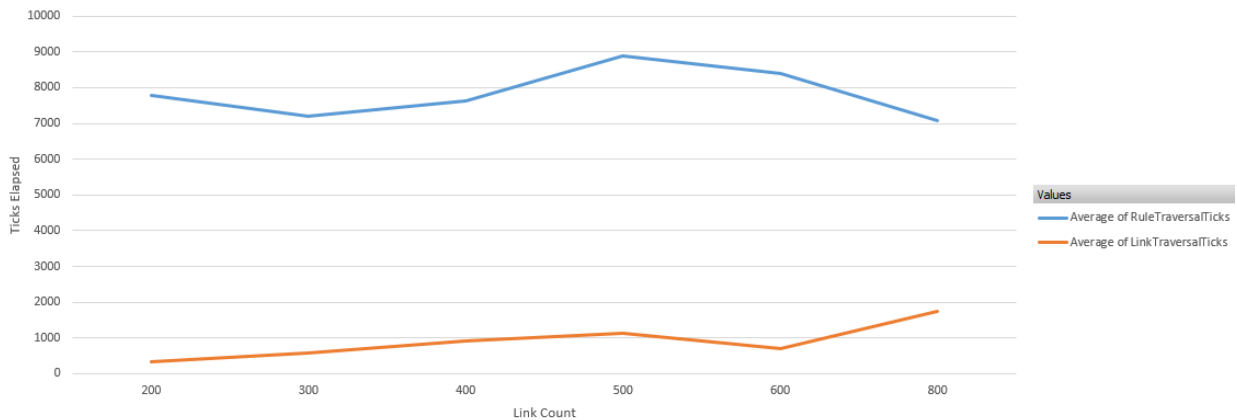

*Figure 14. Link Count vs. Ticks Elapsed*

*Table 10. Data for Link Count vs. Ticks Elapsed*

| Link Count | Average of Rule Traversal Ticks | Average of Link Traversal Ticks |
|---|---:|---:|
| **200** | 7791.69 | 324.31 |
| **300** | 7205.41 | 575.77 |
| **400** | 7622.248 | 904.13 |
| **500** | 8892.48 | 1130.83 |
| **600** | 8401.4 | 693.78 |
| **800** | 7084.84 | 1732.65 |

Figure 14 indicates results that parallel those of Figure 12. The rule traversal time increased, with the increased number of rules. Figure 14 also shows a slight increase in the traversal time for links, as the link count increases. Again, with more links to traverse comes higher traversal times.

Links are another way to organize a network. In the same way that rules create connections between facts, links connect containers. The benefits that links have for traversal come from there being less links than rules that need to be traversed, to get from one fact to another. Because of this, if there are more links than rules in a network it will most likely take longer to traverse via links than via rules. However, for many applications, this will not be the case. Therefore, creating more direct connections between containers with links will often result in faster traversal times than connecting them indirectly via rules.

## 7. Conclusions and Future Work

The use of containers and links within a network helps to bridge the gap between the complexity of Blackboard Architecture networks and the usability that many applications may require. The ability to add layers of organization to a network provides the ability to have different views of any given Blackboard Architecture network. These views can be used, in code, to create customized traversals, connections, and relationships that were unachievable with previous Blackboard Architecture models. The additional organization that is provided by containers and links allows new problems to be conceptualized and analyzed within a Blackboard Architecture network. It also facilitates the effective implementation of more complex networks. The fundamental concept of Blackboard Architecture networks, with rules altering the state of facts, provides a strong foundational representation of many existing structures, processes, and situations. With containers and links, new scenarios can be implemented in an understandable way that provides human-understandable and customizable data that can be used as needed by the system operator.

Several additions could further enhance the capabilities discussed herein. In their current implementation, containers only have one attribute, a description. Depending on the use of containers, it may be useful to add additional fields to increase the storage capability and usability of the objects. One option for an implementation of containers could allow for the arbitrary addition of named attributes, which could then be referenced whenever that specific attribute is needed. The downside of the current implementation, using only one description field, is that any information beyond a simple string description for the container either cannot be stored or must be parsed separately out of the description string. Faster methods of querying attributes could easily be attained by adding a list of attributes to a container. The same concept could be applied to links.

These additional attributes could also be used for traversal of a network. It is possible that a link between two containers could represent a generic logical connection, while a separate attribute defines the traversal behavior over that link. When one is traversing through the network, that specific link attribute is checked, and its value will determine whether or not the link can be traversed.

Many network types provide weightings for their connections between nodes. Links could prospectively support this functionality as well. As with adding directionality, adding weightings to links could provide more detail and functionality that could make designs more realistic and able to be better tailored to the real-life entities that they are created to represent. Weightings could be used to represent several different concepts. One option for weights would be to represent the cost of traversing from one container to another. Another option is to have the weight quantify the level of influence that the start container has on the end container. These weights could be used for traversal or simply be an additional attribute that links store for other use.

In the current implementation, containers may only hold facts. This means that the structure of a network is limited to one abstraction above the level of individual facts. However, there are many scenarios where further abstractions would be beneficial to the design of the network. By allowing containers to hold other containers, the relationships between these abstractions can become layered. This can allow facts to become more organized that they could be with only one layer of containers. This would extend the organization of Blackboard Architecture networks to a much greater level, both hierarchically and relationally. Different layers of containers and subcontainers could be linked, which could create many new types of relationships and organizational levels.

**Acknowledgements**

This work has been funded by the U.S. Missile Defense Agency (contract # HQ0860-22-C-6003).